\documentclass[11pt]{article}

\usepackage[a4paper, margin=1in]{geometry}

\usepackage{comment}
\usepackage{natbib}
\usepackage{booktabs}
\usepackage{graphicx}
\usepackage{amsmath}
\usepackage{amssymb}

\begin{document}
\title[DTI-GP: Bayesian operations for drug-target interactions using deep kernel Gaussian processes]{DTI-GP: Bayesian operations for drug-target interactions using deep kernel Gaussian processes}
\author{
Bence Bolgár\textsuperscript{1} \and
András Millinghoffer\textsuperscript{1} \and
Péter Antal\textsuperscript{1,*}
\\[0.5em]
\textsuperscript{1}Department of Measurement and Information Systems,\\
Budapest University of Technology and Economics, Budapest, 1521, Hungary\\
\textsuperscript{*}To whom correspondence should be addressed
}

\abstract{Precise probabilistic information about drug-target interaction (DTI) predictions is vital for understanding limitations and boosting predictive performance. Gaussian processes (GP) offer a scalable framework to integrate state-of-the-art DTI representations and Bayesian inference, enabling  novel operations, such as Bayesian classification with rejection, top-$K$ selection, and ranking. We propose a deep kernel learning-based GP architecture (DTI-GP), which incorporates a combined neural embedding module for chemical compounds and protein targets, and a GP module. The workflow continues with sampling from the predictive distribution to estimate a Bayesian precedence matrix, which is used in fast and accurate selection and ranking operations. DTI-GP outperforms state-of-the-art solutions, and it allows (1) the construction of a Bayesian accuracy-confidence enrichment score, (2) rejection schemes for improved enrichment, and (3) estimation and search for top-$K$ selections and ranking with high expected utility. \\
}

\maketitle

\section{Introduction}

The field of \textit{in silico} drug discovery, especially the identification of drug-target interactions (DTI) has undergone rapid developments over the past decade. The appearance of high-quality, industry-scale benchmark data sets~\citep{tang2014making,sun2017excape,zhang2024benchmark}, heterogeneous information sources ranging from imagery data~\citep{simm2018repurposing} to genomic screening~\citep{qiu2020bayesian}, and a wide range of machine learning methods~\citep{bagherian2021machine} are all hallmarks of this development, which is also boosted by results in related areas, such as in \textit{de novo} molecule generation~\citep{zhang2022novo}, representation learning for natural products~\citep{atanasov2021natural}, multi-omic fusion~\citep{ye2021unified} and drug repositioning~\citep{arany2013multi}.

However, the performance of DTI prediction is still lagging behind expectations and estimates of future potentials~\citep{heyndrickx2022conformal}, especially in realistic statistical evaluations~\citep{simm21}. Improvements are hindered by data quality~\citep{tang2014making}, the missing-not-at-random property of DTI data~\citep{bolgar2016bayesian}, challenges of real-world active learning scenarios~\citep{warmuth2003active,schneider2018automating}, and the still underutilized transfer effect in multitask DTI learning ~\citep{unterthiner2014multi,xu2017demystifying,wang2022profiling,tian2025enhancing}.

Given the complexity of the pharmaceutical use of DTI prediction, we aim to provide precise probabilistic information beyond binary classification and predicted means for class probabilities and bioactivities. The Bayesian approach offers a principled framework for prior incorporation, model averaging, and reliable quantification of predictive uncertainty; adapting this viewpoint to the DTI prediction problem enables improved operations useful in realistic scenarios, such as filtering, ranking, or prioritization~\citep{antal2003bayesian,moreau2012computational,simm2017macau}. In particular, we formulate our goals as follows: (1) Bayesian inference of means, variances, and highest probability density regions (HPDs), (2) Bayesian classification with rejection, (3) efficient top-$K$ selection, and (4) Bayesian prioritization. This approach adopts the early discovery context focusing on the enrichment of positive hits in the selections and provides quantitative measures for corresponding study design, e.g., distribution of false discovery rate in a top-$K$ selection. Note that the primary focus of our approach remains the construction of accurate predictive posteriors, which can be used to induce posteriors related to sets, preferences, and (partial) orderings as a basis, whereas traditional Bayesian ranking methods aim for a general ranking uniformly for all the entities irrespectively of their relevance and do not provide quantitative predictive information ~\citep{chu2005preference,peska2017drug,chau2022learning}. 

We evaluate the proposed method using the KIBA data set~\citep{tang2014making}. We investigate the performance of the Bayesian DTI-GP approach compared to its \textit{maximum a posteriori} (MAP) variant (DTI-GP-MAP), its extension with rejection; and external references such as DeepDTA~\citep{ozturk18} and SparseChem~\citep{arany2022sparsechem}. We adopt a scaffold-preserving train/test split to avoid over-optimistic performance estimations~\citep{simm21}. 

\section{Materials and methods}
\label{s:methods}

\subsection{Method overview}

Gaussian processes (GP) have been shown to provide accurate estimations of predictive uncertainty in many scenarios and thus are popular candidates for Bayesian inference~\citep{ober21}. This is especially true for their stochastic approximative versions, which fit nicely into current deep learning frameworks and powerful computational hardware, enabling efficient implementations~\citep{gardner18}. In practice, Bayesian inference about DTIs using GP sampling allows (1) the estimation of the distribution of the false discovery rates (FDR) measured on candidate sets, \textit{e.g.} the posterior probability (confidence) that the FDR is above a specified accuracy threshold; (2) rejection schemes to increase enrichment, and (3) estimation of posteriors for top-$K$ selections and ranking.

In this work, we utilize an efficient deep kernel learning (DKL) method~\citep{wilson16}, which combines the flexibility of neural networks and uncertainty estimates of GPs.  The proposed architecture, DTI-GP, incorporates neural networks to construct embeddings from standard ECFP compound fingerprints and precomputed deep protein sequence embeddings and uses these representations as a basis for covariance computations in a stochastic variational GP model.

Sampling from the predictive distribution allows us to estimate posteriors for rankings and top-$K$ candidates~\citep{nguyen2021top}. Since searching for optimal rankings and top-$K$ sets is prohibitive as the number of DTIs can be in the range of $10^{10}\times 10^4$~\citep{heyndrickx2022conformal}, we developed pairwise precedence-based methods, which remain tractable in real-world scenarios and can cope with the inconsistencies of the estimated Bayesian precedence matrix~\citep{brunelli2018survey}. The architecture and workflow are summarized in Figure~\ref{fig:gp-model}. 

\subsection{Data set and preparation}

The KiBA data set is a high-quality kinase data set containing a special aggregated bioactivity score for $467$ targets and $52498$ KiBA compounds~\citep{tang2014making}. Of the $52498$ KiBA compounds, those lacking a smiles descriptor were discarded,
thus $52078$ was retained. SMILES descriptors were then standardized using the
software tool RDKit, with the following results: $83$ compounds, which contained
more than 100 non-H atoms were discarded; among the remaining $51995$ canonical
smiles descriptors, there were $50418$ unique ones, which then formed the final
list of compounds, having merged the activation data corresponding to the
original smiles descriptors.

Since completely random train/test splits suffer from the compound series bias, leading to overoptimistic performance estimations~\citep{mayr18}, we utilize a more realistic, scaffold-based train/test split in the spirit of~\citep{simm21}.
The scaffold to be assigned to the compound was selected in two steps: first, the
Murcko scaffold belonging to the compound was retrieved using RDKit; second, this
Murcko scaffold was used to select the preferred scaffold from the tree of
scaffolds. Thus, the $50418$ unique canonical smiles descriptors were assigned to
$9382$ scaffolds. Finally, a random number between $0$ and $5$ was generated for each of the above
scaffolds, assigning the corresponding compounds to the given fold (folds are available at the repository). 

Following DeepDTA~\citep{ozturk18}, we use a binary discretization scheme with a threshold of $pK_d=3$, resulting in $72944$ and $162681$ active and inactive DTIs, respectively. The data are summarized in Table~\ref{tab:KIBA}).

\begin{table}[!t]
\caption{Description of the dataset.}
\label{tab:KIBA}
\centering
\begin{tabular}{@{}llllll@{}}
\toprule
Compounds & Proteins & Interactions & Active & Inactive & Sparsity \\
\midrule
50418 & 467 & 235625 & 72944 & 162681 & 0.01 \\
\bottomrule
\end{tabular}
\end{table}

\subsection{Proposed model}

Despite being ``somewhat'' Bayesian, DKL-based GP models are prone to overfitting due to their tendency to overcorrelate input data. This occurs essentially because the covariance function is still learned through maximum likelihood estimation, which is not a problem with only a handful of kernel parameters but can be detrimental to generalization performance when the covariance function is learned by a highly overparameterized deep neural network~\citep{ober21}. To mitigate this problem, we delegate much of the representation learning to the problem-agnostic preprocessing phase and use a stochastic variational approximation that provides implicit regularization through minibatching.

\begin{figure}[!tpb]
\centerline{\includegraphics[width=\linewidth]{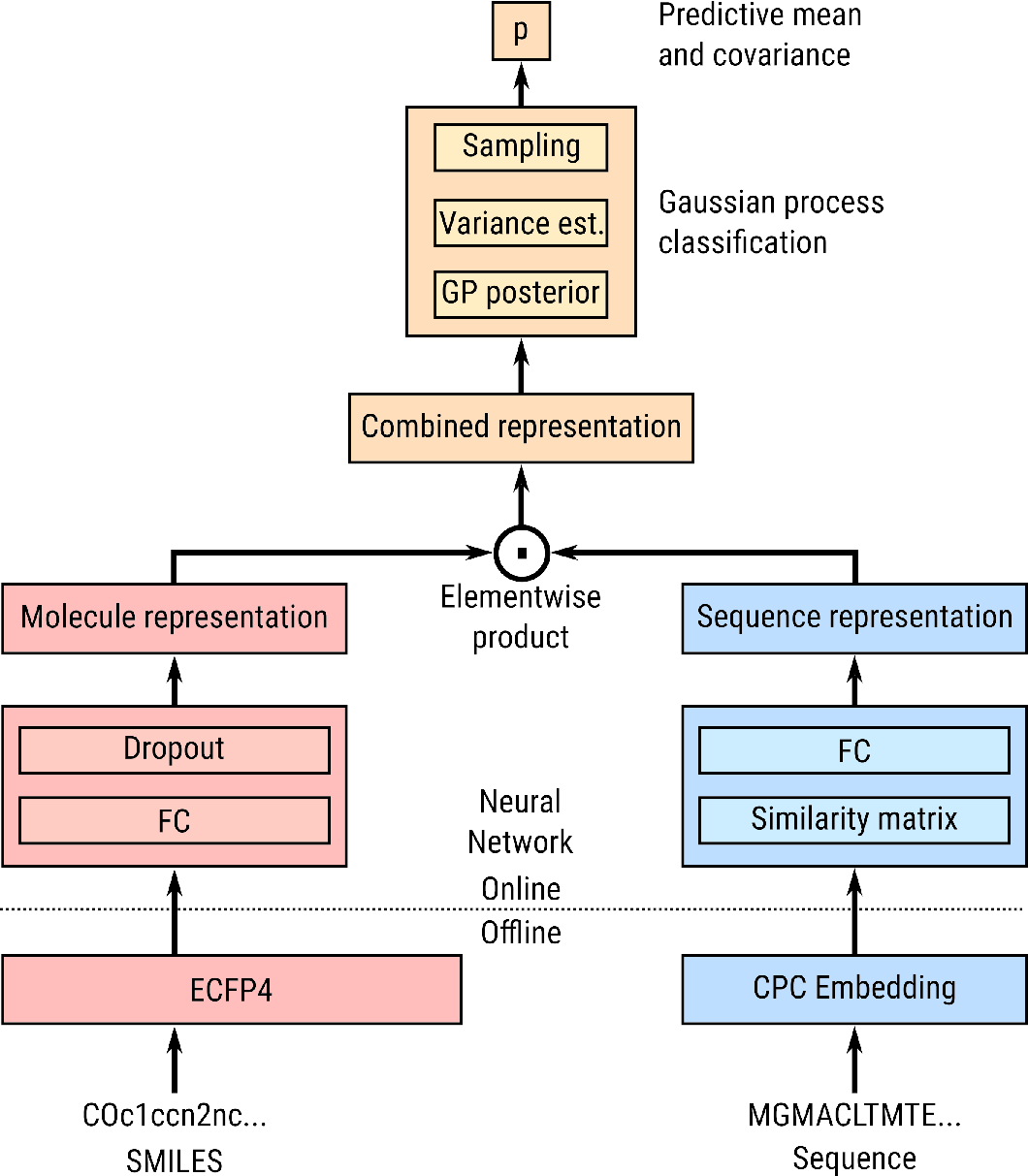}}
\caption{DTI-GP model with two neural networks learning molecule and sequence representations and a Gaussian process classification model computing probabilistic outputs.}
\label{fig:gp-model}
\end{figure}

\subsubsection{Input representations}

Contrary to earlier works, we do not use a deep neural network to learn representations for molecules; instead, we utilize the RDKit package to compute ECFP4 fingerprints of length $32$k from SMILES strings. Similarly, protein representations are calculated offline using the bio-embeddings Python package~\citep{dallago21}. We experimented with several protein embeddings and found $512$-dimensional CPCProt embeddings~\citep{lu20} lead to top predictive performance.

\subsubsection{Learning the covariance function}

To minimize overfitting, we utilize a neural architecture with a relatively small size. In particular, we use a neural network with only one hidden layer with $512$ neurons for the molecules. For the proteins, we compute an RBF similarity matrix with learnable parameters. Subsequently, we compute $512$-dimensional embeddings for both molecules and proteins; embeddings are then combined using elementwise multiplication and serve as inputs for covariance computation. Formally, we construct combined pairwise representations
\begin{align*}
\mathbf x_{\text{pw}} = NN_{\text{mol}} (\mathbf x_{\text{FP}}) \odot NN_{\text{prot}}(\mathbf S(x_\text{CPC})),
\end{align*}
where $\mathbf x_{\text{pw}} \in \mathbb{R}^{512}$ is the combined embedding, $\mathbf x_{\text{FP}} \in \mathbb{R}^{32000}$ is the ECFP4 fingerprint, $\mathbf{x}_\text{CPC} \in \mathbb{R}^{512}$ is the pre-computed protein embedding, $\mathbf S$ is the similarity function, $NN_{\text{mol}}$ and $NN_{\text{prot}}$ are the molecule and protein encoders.

\subsubsection{Bayesian prediction through Gaussian processes}

Gaussian processes (GP) are often preferred when reliable uncertainty estimates are needed. The goal of GP methods is to learn a function $f$, which takes the $i$th embedding $\mathbf{x}_{\text{pw},i}$ to the corresponding label $y_i$:
\begin{align*}
y_i \approx f(\mathbf{x}_{\text{pw},i}).
\end{align*}
Let $\mathbf{f}$ denote the values of $f$ when applied to all embeddings $\mathbf{X}_{\text{pw}} = \left\lbrace \mathbf{x}_{\text{pw},i} \right\rbrace_{i=0}^N$, where $N$ is the number of interactions in the training set. Thinking of $\mathbf f$ as a probabilistic object, we have a prior
\begin{align*}
p\left(\mathbf{f} \right) = \mathcal{N} \left( \mathbf{f} \mid \mathbf{m}, \mathbf{K}\right),
\end{align*}
where we choose the mean function $\mathbf{m}$ to be a learnable constant, and $\mathbf{K}$ is the covariance matrix, which is computed via a kernel function $k$, \textit{i.e.}
\begin{align*}
\mathbf{K}_{ij} = k\left(\mathbf{x}_{\text{pw},i},\mathbf{x}_{\text{pw},j}\right) \quad\text{or}\quad \mathbf{K} = k\left(\mathbf{X}_{\text{pw}},\mathbf{X}_{\text{pw}}\right) .
\end{align*}

In the classification setting, the likelihood over the labels $\left\lbrace y_i \right\rbrace_{i=0}^N$ is Bernoulli:
\begin{align*}
p\left( \mathbf{y} \mid \mathbf{f}\right) = \prod_{i=1}^N \mathcal{B}ern \left( y_i \mid \Phi(f(\mathbf x_{\text{pw},i})) \right),
\end{align*}
where $\Phi$ is the Gaussian CDF function.

This model is non-conjugate, \textit{i.e.} the posterior and predictive distributions cannot be calculated in a closed form. We utilize stochastic variational Gaussian process classification of~\citep{hensman14}, an approach that has been shown to provide well-calibrated outputs without resorting to post-hoc calibration when combined with deep kernel learning and appropriate minibatching~\citep{ober21}. In particular, to approximate the posterior $p\left(\mathbf f \mid \mathbf{y} \right)$, we specify a variational distribution $q(\mathbf f)$ and minimize the KL-divergence $KL\left( q(\mathbf f) \,||\, p\left(\mathbf f \mid \mathbf{y} \right) \right)$.

The variational distribution depends on a small number of pseudo-inputs and can be specified as
\begin{align*}
q(\mathbf f) = \mathcal{N} \left( \mathbf f \,\Big\vert\, \mathbf A \mathbf \mu, \mathbf A \left(\mathbf \Sigma - \mathbf{K}_{uu} \right) \mathbf A^\top \right),
\end{align*}
where $\mathbf{A} = \mathbf{K}_{fu} \mathbf{K}^{-1}_{uu} $, $\mathbf{K}_{fu}$ is the kernel matrix computed between the real inputs and pseudo-inputs, $\mathbf{K}_{uu}$ is the kernel matrix computed between the pseudo-inputs, and $\mathbf{\mu}$, $\mathbf{\Sigma}$ are the variational parameters, \textit{i.e.}
\begin{align*}
q(\mathbf u) = \mathcal{N}\left( \mathbf u \mid \mathbf{\mu}, \mathbf{\Sigma}\right).
\end{align*}
Since the linear solves only need the kernel over the pseudo-inputs $\mathbf{K}_{uu}$, the optimization of the variational parameters remains tractable. Instead of directly minimizing the KL divergence, we use the GPyTorch library's fast CG-based solver to compute the evidence lower bound
\begin{align*}
\mathbb{E}_{q(\mathbf f)} \left[ p\left( \mathbf y \mid \mathbf f \right) \right] - KL\left(q(\mathbf{u}) \,||\, p(\mathbf{u}) \right),
\end{align*}
which is readily optimized using the solvers from PyTorch, together with the neural network parameters.

\subsection{Bayesian top-$K$ selection with sampling}

The predictive distribution can be efficiently sampled using Lanczos variance estimates~\citep{pleiss18}. We use the collected samples to establish a pairwise precedence matrix $\mathbf{P}$, where $\mathbf P_{ij} = p(\mathbf{f}^*_i > \mathbf{f}^*_j)$ for a test set $\mathbf{X}_{\text{pw}}^*$. We think of $\mathbf{P}$ as a generalized \emph{tournament matrix} and propose two heuristics for selecting top-$K$ interactions:
\begin{itemize}
\item {\bf Score.} Compute the normalized score for each interaction in the test set, \textit{i.e.} $\frac{1}{N^*}\sum_j \mathbf{P}_{ij}$ and select test interactions with the top $K$ scores. Intuitively, this selects interactions that have consistently larger predicted probabilities than others.
\item {\bf Eigen.} Compute the dominant eigenvector of $\mathbf{P}$ via power iteration and use it to select top $K$ interactions. Due to power iteration, this can be interpreted as a multi-step generalization of the former, taking all ordered paths into account.
\end{itemize}
Similarly to the reliability diagram, Figure~\ref{fig:precedence} depicts the relationship between the binned values and the real fraction of positives in each bin.

\subsection{Baseline methods}

We compare predictive performance to a state-of-the-art model, DeepDTA, a deep neural network-based model using compound and protein sequence data for predicting binding affinities. DeepDTA utilizes convolution blocks in the compound and protein encoders, forms a combined representation by concatenation, and utilizes fully connected layers in the predictor network. Although DeepDTA was proposed for affinity prediction (\textit{i.e.} continuous values), it is also evaluated in a classification framework in the original publication, yielding top AUPR values~\citep{ozturk18}.

We also compare predictive performance to the SparseChem model~\citep{arany2022sparsechem}, which bears the closest resemblance to the proposed architecture. SparseChem utilizes a shallow MLP with sparse inputs (ECFP fingerprints) and sparse multitask outputs to predict binary and continuous bioactivity values.

\section{Experiments and results}

We ran the experiments on a machine with a 32GB NVIDIA Tesla V100 GPU. We utilize the area under the ROC curve and precision-recall curve as performance metrics. As a baseline, we use the code from the DeepDTA authors' GitHub repository with the recommended hyperparameters, built-in hyperparameter search, and regression-on-classification-labels approach. Exploiting the Bayesian nature of the proposed model, we investigate the rejection option, where we evaluate using only the values in the test set where the model is fairly confident ($\text{std} < 0.05$). We also evaluate predictive performance when the Bayesian approach is omitted, \textit{i.e.} using only MAP estimation by choosing the variational distribution as a Dirac delta.

We evaluated the predictive performance in a global and a protein-wise setting. More specifically, the global evaluation does not take the columns of the interaction matrix into account and uses ROC and PR curves on the whole test set. In the protein-wise setting, we compute AUROC and AUPR values on every column of the interaction matrix where the corresponding protein has at least $50$ positive and $50$ negative interactions.

\begin{figure}[!tpb]
\centerline{\includegraphics[width=\linewidth]{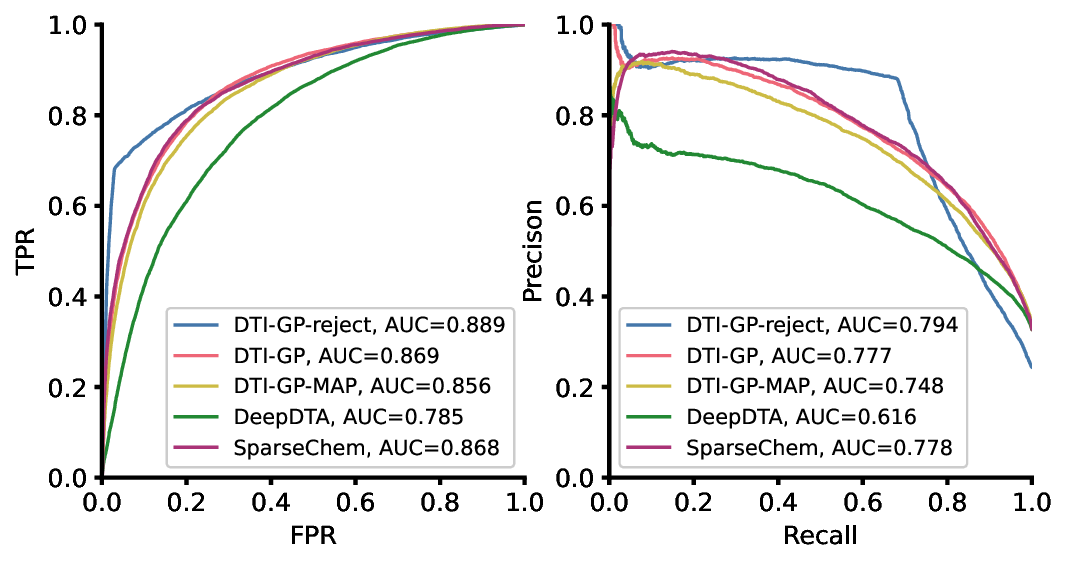}}
\caption{ROC and PR curves in the global evaluation setting. The GP-based model outperforms DeepDTA, especially in the region relevant for early discovery and top-$K$ selection.}
\label{fig:auroc_aupr}
\end{figure}

Figure~\ref{fig:auroc_aupr} presents the results of the global evaluation. It is worth noting that the GP-based approach outperforms DeepDTA and SparseChem in the low recall/high precision domain, which is especially relevant from the early discovery and top-$K$ selection viewpoint. The MAP estimation, although outperforming DeepDTA, yields more modest results with poor performance in the top-$K$ selection region.  Table~\ref{tab:protwise} shows the results in the protein-wise evaluation.

\begin{table}[!t]
\caption{Predictive performance in a task-wise evaluation.}
\label{tab:protwise}
\centering
\begin{tabular}{@{}lll@{}}
\toprule
Method & AUROC & AUPR \\
\midrule
DTI-GP-reject & 0.881 $\pm$ 0.086 & 0.833 $\pm$ 0.117 \\
DTI-GP        & 0.843 $\pm$ 0.081 & 0.801 $\pm$ 0.092 \\
DTI-GP-MAP    & 0.834 $\pm$ 0.090 & 0.783 $\pm$ 0.099 \\
SparseChem    & 0.851 $\pm$ 0.091 & 0.816 $\pm$ 0.092 \\
DeepDTA       & 0.751 $\pm$ 0.064 & 0.666 $\pm$ 0.099 \\
\bottomrule
\end{tabular}
\end{table}

\begin{figure}[!tpb]
\centerline{\includegraphics[width=\linewidth]{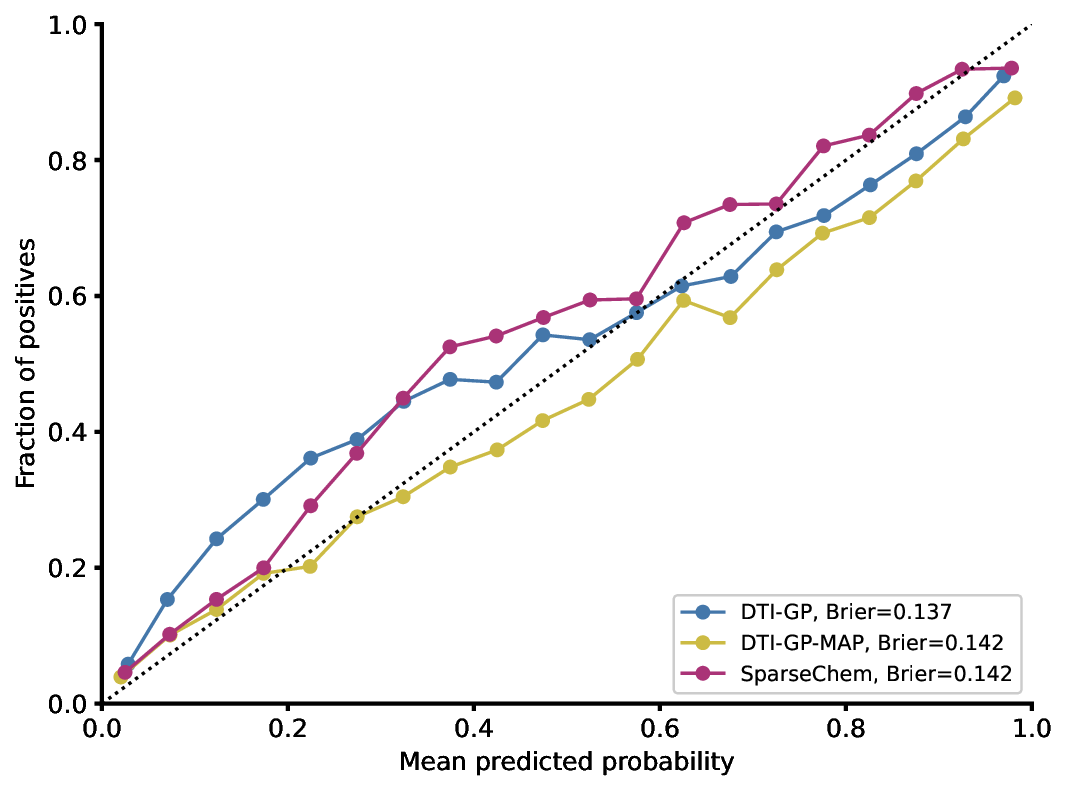}}
\caption{Reliability plot of the proposed model. Note that both the Bayesian and MAP estimations suffer from overconfidence in the relevant high-probability region.}
\label{fig:calibration}
\end{figure}

We examined the calibration properties of the model and found that despite considerable effort, both the Bayesian and the MAP versions of the model are overconfident in the high-probability region (Fig.~\ref{fig:calibration}). This behavior remains even after applying Monte Carlo dropout, which has been found to improve calibration~\citep{tran19}. The MAP estimation especially suffers from this type of miscalibration, and as the next Section shows, this anomaly profoundly impacts the ability to choose a good top-$K$ set. SparseChem also shares this trait, although generally underconfident in other regions, which makes its top-$K$ selection performance even worse.

\begin{figure}[!tpb]
\centerline{\includegraphics[width=\linewidth]{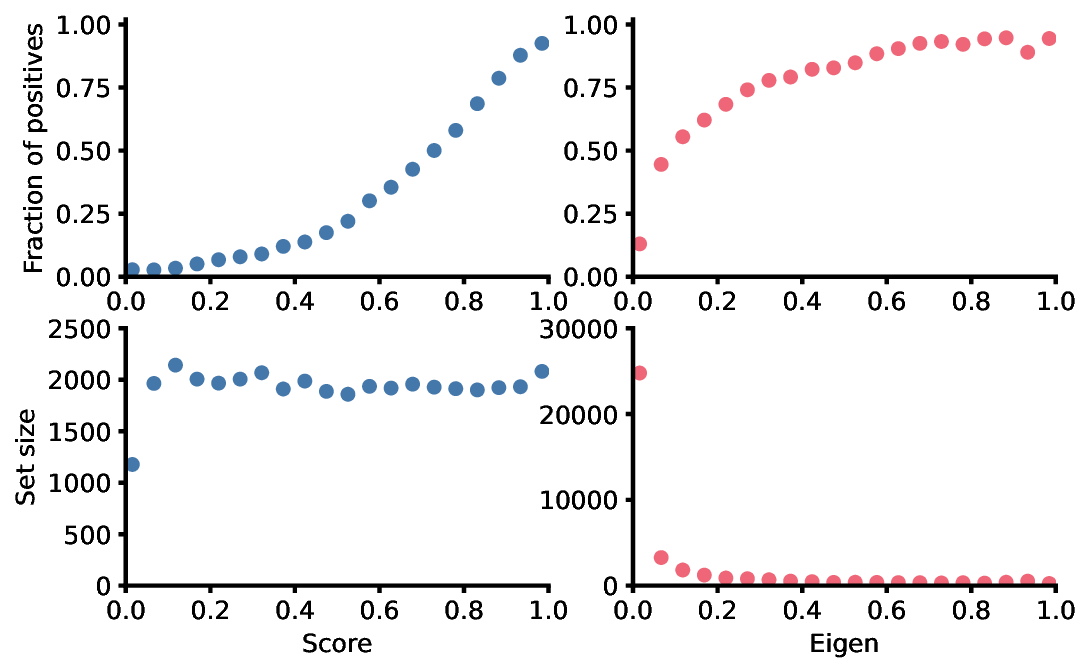}}
\caption{Difference between the score and eigenvector-based heuristics. The eigenvector-based method assigns values close to $0$ to most interactions; the score-based method distributes values more evenly.}
\label{fig:precedence}
\end{figure}

\begin{figure}[!tpb]
\centerline{\includegraphics[width=\linewidth]{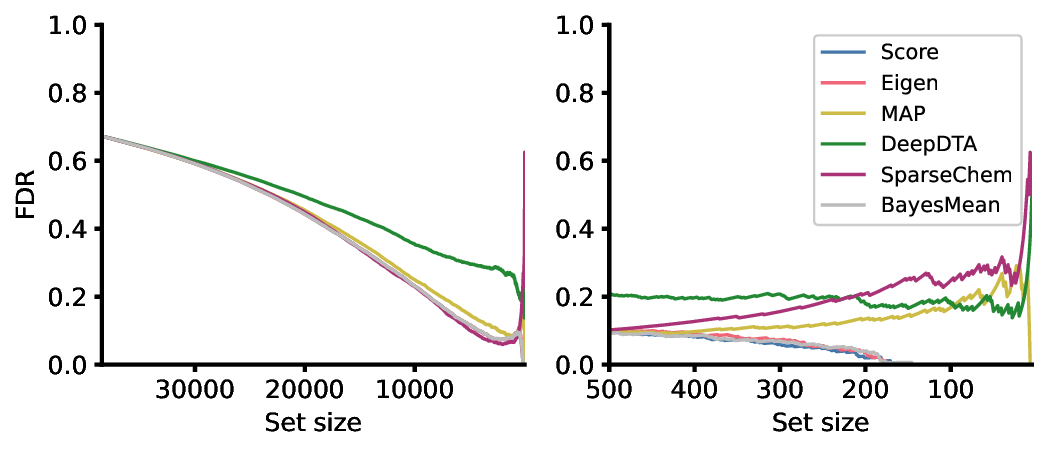}}
\caption{Relationship between the set size $K$ and false discovery rates. The score and eigenvector-based heuristics outperform DeepDTA, SparseChem and DTI-GP-MAP in the small set size region. The right side of the figure is somewhat noisy, as expected with such small set sizes.}
\label{fig:setsize_fdr}
\end{figure}

To assess the top-$K$ sets constructed using the heuristics, we investigate the relationship between the set size $K$ and false discovery rates (Figure~\ref{fig:setsize_fdr}). Both proposed heuristics construct almost identical sets in terms of FDR, and both outperform the top-$K$ sets constructed from the class probabilities predicted by DeepDTA and MAP estimation. The difference is more pronounced in the small set size region, which is more important from a practical viewpoint. For large set sizes, SparseChem performs almost identically to the proposed heuristics, but for smaller sets, performance gradually deteriorates. This underlines the advantage of using Bayesian averaging and precedence-based methods as compared to maximum likelihood and MAP-type point estimates.

\section{Discussion}

The multitask learning approach to DTI prediction holds the promise of a transfer effect between related protein targets or even assays. Pairwise DTI approaches using dual representations for compounds and targets reached the performance of DTI predictors with fixed tasks~\citep{ozturk18}. However, in a real-world pharmaceutical scenario, uniformly good quantitative estimates for all DTIs, class probabilities or expected bioactivity values are not necessary. On the other hand, precise quantitative, probabilistic information about the utility of an experiment in early drug discovery is vital, \textit{e.g.}, to explore the distribution of the FDR for a candidate set in the range of a thousand DTI for high-throughput screening or a dozen DTIs for high precision assaying. To exploit the robustness of the underlying idea of learning to rank approaches and still providing quantitative probabilistic information, we proposed a method with dual Bayesian averaging: the proposed DTI-GP model performs Bayesian model averaging in the GP framework, and the DTI-GP predictions are used to induce a Bayesian precedence matrix to collate the voluminous, optionally weak ranking information between individual DTIs. 

\begin{figure}[!tpb]
\centerline{\includegraphics[width=\linewidth]{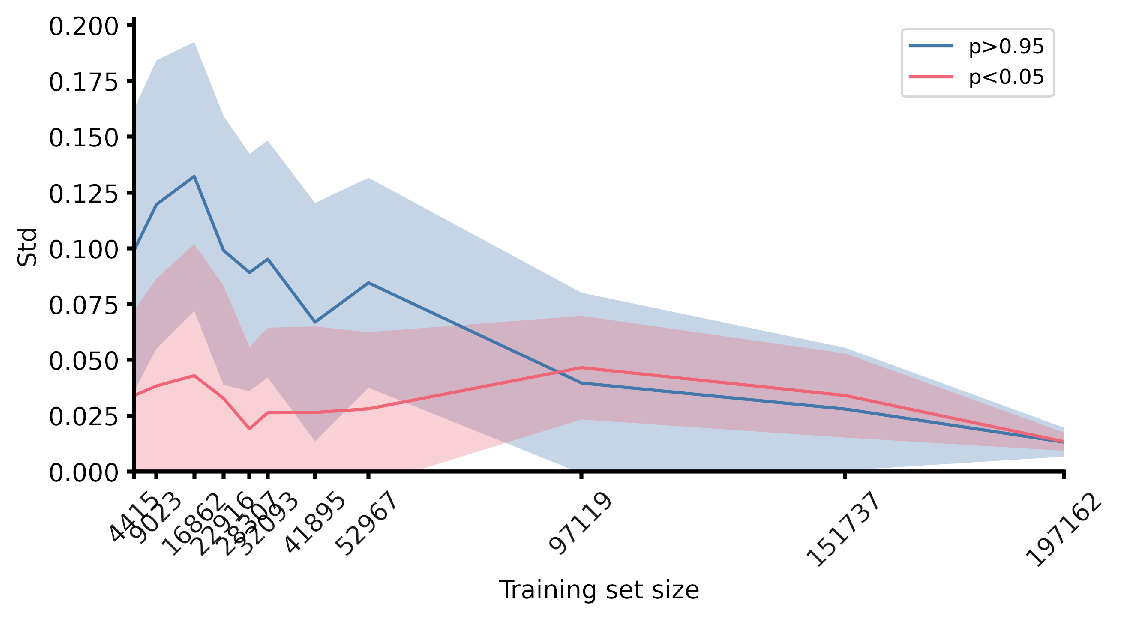}}
\caption{Standard deviation measured on the top and bottom part of the test set, with the parts established according to the predictions using the full training set.}
\label{fig:std_curve}
\end{figure}

The DTI-GP model demonstrates that Bayesian pairwise DTI prediction (see Figure~\ref{fig:gp-model}) achieves top predictive performance (see Figure~\ref{fig:auroc_aupr} and Table~\ref{tab:protwise}). This confirms that pairwise architectures, applicable for any represented task compared to DTI predictors with fixed multiple task outputs are competitive. DTI-GP provides calibrated \textit{a posteriori} class probabilities (see Figure~\ref{fig:calibration}). In addition, Bayesian approaches can provide a multivariate posterior distribution over the predictive class probabilities quantifying their marginal and overall uncertainty. Figure~\ref{fig:std_curve} shows the learning curves for the  estimated variances for the predicted class posteriors. Based on all training data, we formed two groups for the expectedly inactive and active DTIs (respectively, when the predicted posterior of an active DTI $p$ is $p<0.05$ and  $0.95<p$). As early discovery uses only top estimates, it is worth noting that variances for top candidates follow different characteristics and decrease at different rates.   

\begin{figure}[!tpb]
\centerline{\includegraphics[width=\linewidth]{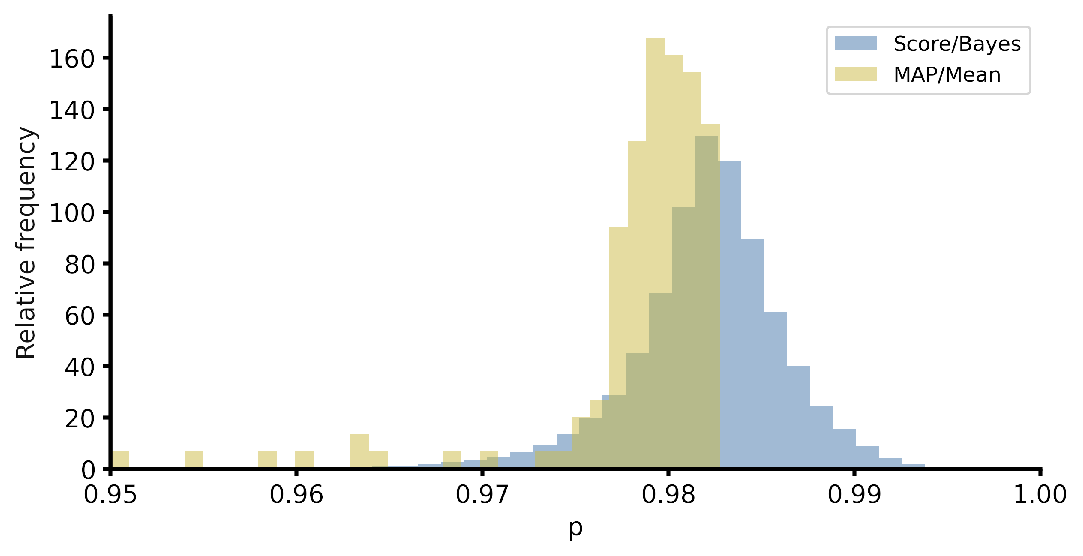}}
\caption{Histograms of the predicted class probabilities using the top-$K$ sets ($K=150$) from the Score- and MAP-based selection.}
\label{fig:top150_density}
\end{figure}

\begin{figure}[!tpb]
\centerline{\includegraphics[width=\linewidth]{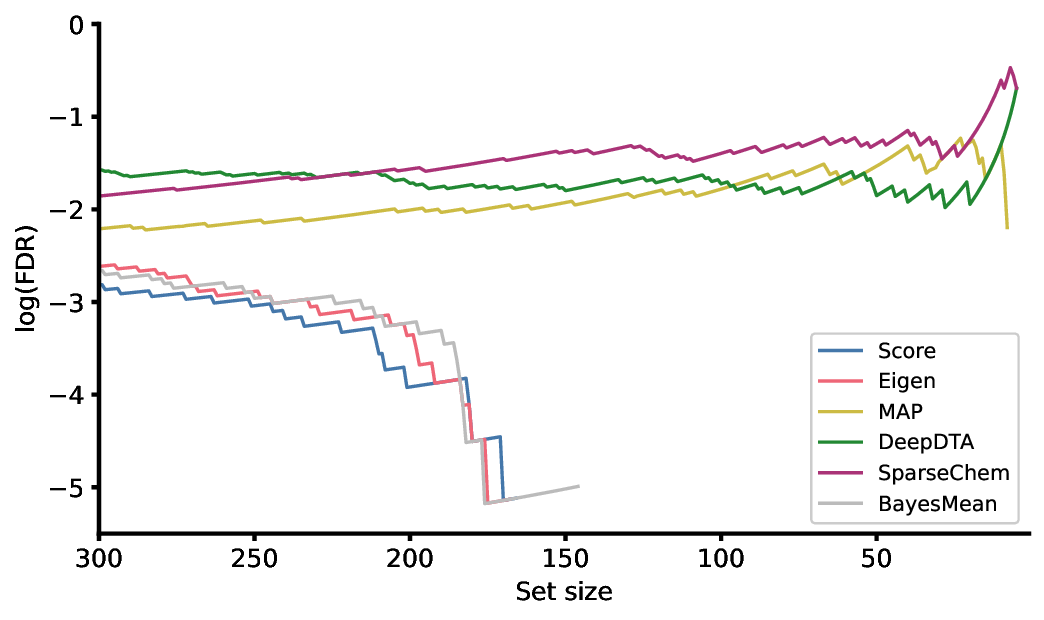}}
\caption{Relationship between the set size $K$ and false discovery rates in the small set region. The Score, the eigenvector-based heuristics (Eigen) and the DTI-GP mean estimate based top-$K$ selections (BayesMean) far outperform DeepDTA, SparseChem, and the MAP version of the model in the small set size region.}
\label{fig:zoom}
\end{figure}

The DTI-GP workflow heavily utilizes this Bayesian inference about DTIs: using a fast GP sampling method, we estimate a Bayesian precedence matrix over DTIs, which is meant to capture high-level ranking information averaging over the quantitative levels. Note that predicted point-values for class posteriors always offer a consistent global ranking (except for ties), but the performance of top-$K$ selection methods based on this global ranking is inferior compared to Bayesian precedence-based methods (see Figure~\ref{fig:precedence} and~\ref{fig:setsize_fdr}). Figure~\ref{fig:top150_density} compares the performance of the MAP approximation versus a full/analytic Bayesian model averaging in the GP framework: it shows the histogram of the GP sampled class posteriors for the top-$K$ DTIs for $K=150$ selected by the Score method and the corresponding histogram of the mean class posteriors for the top-$K$ DTIs determined by DTI-GP-MAP (\textit{i.e.} the selection is from DTI-GP-MAP, but the mean values are from the more precise full Bayesian DTI-GP). As we can see, predicted mean posteriors for candidate DTIs spread to a wide range, and there is no candidate DTI with a predicted mean posterior above $0.985$ from the DTI-GP-MAP. Whereas the full Bayesian DTI-GP predicts with more than $50\%$ confidence that there are hits with higher posteriors than $0.985$, and their spread is more concentrated. Such information can influence the design and decision about a screening program.  
The performance of DTI-GP based enrichment is especially striking in the small candidate set region ($<500$). Figure~\ref{fig:zoom} shows an enlarged part of this small candidate set region.

The inherent inference capability of GPs to perform Bayesian averaging over the covariances achieves the same level of enrichment performance in top-$K$ selection as the Bayesian precedence-based solutions (see Score and Eigen versus BayesMean in Figure~\ref{fig:zoom}). However, note that the Bayesian precedence matrix can support a wide range of operations beyond top-$K$ selection, such as inference of partial ranking or ranking of hierarchies which we plan to explore in the future.

\section{Conclusion}

Results confirm that the more flexible pairwise DTI predictors using explicit target representations together with compound representations can achieve the same performance as the more restricted architectures with a fixed set of targets. This suggests that more challenging entities, such as natural products instead of compounds and assays instead of target proteins, could also be tackled in pairwise DTI predictors. A somewhat unexpected result is that shallow architectures can achieve significantly better performance than deep architectures (we explored a wide range of architectural and deep learning options for DTI-GP-MAP). This issue highlights the importance of more detailed benchmarks, specifically with the pharmaceutically motivated scaffold-preserving cross-validation folds~\citep{simm21}.  Precisely calibrated mean predictions and measures to quantify prediction uncertainty are much needed in efficient pharmaceutical study design, \textit{e.g.}, to select a candidate set for screening with prespecified FDR characteristics. DTI-GP provides calibrated mean predictions, but its novel feature is that it also offers efficient distributional information. The proposed Bayesian operations offer unique enrichment performance together with its  probabilistic characterization appropriate for full-fledged decision support, indicating significant advances. A cautionary note is that combinations of the predicted means and variances successfully applied in multi-armed bandits and sequential learning with cumulative losses did not lead to useful rejection and enrichment schemes in the adopted one-step decision framework using only the known DTIs for training and testing. However, we anticipate many possibilities of DTI-GP in the active and reinforcement learning frameworks, \textit{e.g.}, evaluating its performance using unknown DTIs as well. The realized DTI-GP architecture and workflow are scalable and preliminary results confirm these results on industry-scale data set~\citep{sun2017excape}.


\section*{Funding}

This work has been supported by the ÚNKP-21-4 New National Excellence Program of the Ministry for Innovation and Technology from the source of the National Research, Development and Innovation Fund, OTKA-K139330, the European Union project RRF-2.3.1-21-2022-00004 within the framework of the Artificial Intelligence National Laboratory. This research was carried out as part of E-GROUP ICT SOFTWARE Zrt.'s “IPCEI-CIS FedEU.ai – Federated Cloud-Edge AI” project, supported by the Government of Hungary and the Ministry of National Economy, under grant number NGM/4043/1/2024. The authors acknowledge the support of E-GROUP ICT SOFTWARE Zrt. for providing the research infrastructure and resources essential to this work. The research presented in this paper forms part of E-GROUP’s core research domain and contributes to the advancement of its strategic directions in federated, cloud-edge, and cutting-edge AI technologies.


\end{document}